\pdfoutput=1
\documentclass[11pt]{article}

\usepackage{ACL2023}

\usepackage{times}
\usepackage{latexsym}
\usepackage{graphicx}
\usepackage{booktabs}
\usepackage{multirow}
\usepackage{amsmath}
\usepackage{colortbl}
\usepackage[T1]{fontenc}
\usepackage[whole]{bxcjkjatype}
\usepackage[utf8]{inputenc}

\usepackage{microtype}

\usepackage{inconsolata}
\usepackage{ulem}

\title{\textsc{JapaGen}: Efficient Few/Zero-shot Learning \\ via Japanese Training Dataset Generation with LLM}
\author{Takuro Fujii$^{\text{1,2,}*}$ \and Satoru Katsumata$^\text{3}$ \\
         $^\text{1}$Yokohama National University \hspace{1mm} $^\text{2}$Nomura Research Institute, Ltd. \hspace{1mm} $^\text{3}$Retrieva, Inc. \\
         \texttt{tkr.fujii.ynu@gmail.com} \hspace{7mm}\texttt{satoru.katsumata@retrieva.jp}\\
         }

\begin{document}
\maketitle
{\let\thefootnote\relax\footnotetext{$^{\text{*}}$ Work done while internship at Retrieva, Inc. when I was a master student. Now I belong to Nomura Research Institute, Ltd.}}
\begin{abstract}

Recently some studies have highlighted the potential of Large Language Models (LLMs) as effective generators of supervised training data, offering advantages such as enhanced inference efficiency and reduced costs associated with data collection. However, these studies have predominantly focused on English language tasks.
In this paper, we address the fundamental research question: \textit{Can LLMs serve as proficient training data generators for other language tasks?} Specifically, we leverage LLMs to synthesize supervised training data under few-shot and zero-shot learning scenarios across six diverse Japanese downstream tasks. Subsequently, we utilize this synthesized data to train compact models ({\textit{e.g.}}, BERT). This novel methodology is termed \textsc{JapaGen}.
Our experimental findings underscore that \textsc{JapaGen} achieves robust performance in classification tasks that necessitate formal text inputs, demonstrating competitive results compared to conventional LLM prompting strategies.

%Large Language Models (LLMs) can solve tasks with high performances even in the case of few/zero shot setup.
%However, it is high cost to use LLMs for inference due to large parameters.
%To approach the problem, some works trained small language model ({\textit{e.g.}}, BERT) with supervised data generated using LLMs in few/zero shot setup.
%These works showed the method enables to keep high performances and decrease costs of data collection, annotation, and inference, all of which were conducted in English tasks.
%In this paper, we investigate how it works well in Japanese tasks: {\textit{JapaGen}}.
%Specifically, we evaluate {\textit{JapaGen}} in six Japanese downstream tasks in both few/zero shot setup.
%Our experimental resutls showed our JapaGen obtained higher performances than few-shot Finetuning or Propmpting in some tasks.
\end{abstract}

\section{Introduction}

% 流れ：
% LLMはいろいろすごいよね。でもコスト高い。LLMで学習データを生成する研究がある。LLMで学習データ生成するメリット（推論コスト、データ収集・アノテーションコスト）。
% LLMより前の時代の学習データ生成のお話。でもこれらはデータ拡張に基づくものであり、もともとのデータセットがないとだめなのでデータ収集コストが高い。
% 先行研究では～～なタスクでPromptingに勝ったり、～～な結果を得ている。しかし、それらは全てhigh-resourceな英語タスク。mid-resourceであり言語特性も異なる（単語間スペース）日本語でも同じようにworkするのか。
% 本研究では、、、を検証する。具体的には、、、。
% 実験の結果
% 貢献

%Large language models (LLMs) have achieved remarkable performance in various natural language processing (NLP) tasks~\cite{opt,holistic,gpt4}, even if no parameter updating~\cite{few-shot_learner,cot}.
%Large language models (LLMs) have achieved remarkable performance in various natural language processing (NLP) tasks even if no parameter updating~\cite{few-shot_learner,cot}. However, with the rapid growth of the parameter size based on scaling laws~\cite{scaling_law}, large GPU memory consumption and massive computation are required. Therefore, it is too expensive to operate LLMs.

Large language models (LLMs) have demonstrated exceptional performance across various natural language processing (NLP) tasks, even with minimal parameter updates~\cite{few-shot_learner,cot}. However, the rapid growth in model size, driven by scaling laws~\cite{scaling_law}, has led to substantial demands for GPU memory and computational resources, making the operation of LLMs prohibitively expensive.

%To reduce such costs, some studies have attempted to generate training data from scratch using a powerful LLMs, and then, train a smaller model ({\textit{e.g.}, BERT}) than LLMs under the synthesized supervised training data~\cite{zerogen,progen,regen,chung-etal-2023-increasing}. We refer a scenario of such an approach to \textsc{SuperGen}, Supervision Generation Approach, based on \cite{supergen}. The overview of \textsc{SuperGen} is shown in Figure \ref{fig:overview}. Those studis demonstrated that \textsc{SuperGen} ourperformed few/zero-shot prompting and few-shot finetuning in some tasks, all of which were conducted within English tasks. \textsc{SuperGen} can reduce costs both to collect supervised data and operate a trained model.

\begin{figure}
\includegraphics[width=\linewidth]{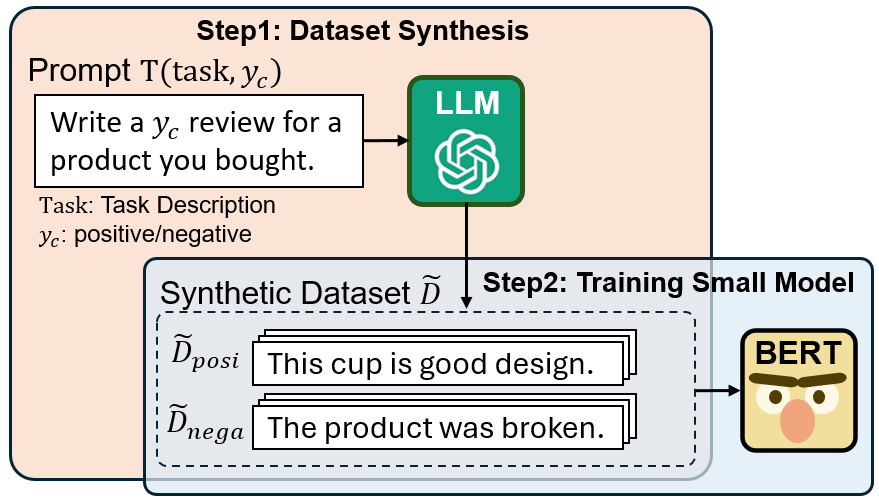}
   \caption{Overview of \textsc{SuperGen} in text sentiment classification as an example.}
\label{fig:overview}
\end{figure}

To mitigate these costs, recent studies have investigated the generation of training data using powerful LLMs, followed by training smaller models (e.g., BERT) on the synthesized supervised data~\cite{zerogen,progen,regen,chung-etal-2023-increasing}. This approach, termed \textsc{SuperGen} (Supervision Generation Approach) based on prior work~\cite{supergen}, has demonstrated promising results. The overview of \textsc{SuperGen} is illustrated in Figure \ref{fig:overview}. \textsc{SuperGen} has been demonstrated to outperform few-shot and zero-shot prompting and few-shot fine-tuning methods in various tasks, effectively reducing both the cost of collecting supervised data and the operational costs of trained models.
However, these studies have been limited to English tasks, and thus, the applicability of \textsc{SuperGen} on other language tasks remain uncertain.

%In this paper, we investigate the research question "\textit{Do SuperGen work well in Japanese?}". We refer \textsc{SuperGen} in Japanese tasks to \textsc{JapaGen}. Japanese is middle-resource language compared to English, and also has different characteristics from English in that no spaces exist between words. Powerful LLMs like GPT-4~\cite{gpt4} are trained mainly with English texts, and a few texts in other language including Japanese. It is interesting whether \textsc{SuperGen} in such languages work well, particularly for what kinds of tasks. 

Given that powerful LLMs like GPT-4~\cite{gpt4} are primarily trained on English texts with limited exposure to other languages, it is crucial to investigate the effectiveness of \textsc{SuperGen} in such linguistic contexts and its suitability for different types of languages.
%In this paper, we treat Japanese as a language of medium-resource language compared to English and unique characteristics such as lack of spaces between words.
In this paper, we implement \textsc{SuperGen} in Japanese as a case study. Japanese is mid-resource language compared to English and has different characteristics, such as the absence of spaces between words.
Therefore, we pose the research question: \textit{Do SuperGen methods perform effectively in Japanese?} We term the application of \textsc{SuperGen} to Japanese tasks as \textsc{JapaGen} (\S\ref{sec:japagen}).
%Japanese represents a medium-resource language compared to English and possesses unique characteristics such as lack of spaces between words.

%To approach aforementioned interests, we evaluate \textsc{JapaGen} in four Japanese tasks including text classification, natural language inference, semantic textual similarity, linguistic acceptability across six tasks. Unlike previous works, we evaluate \textsc{JapaGen} on both few- and zero-shot setup. As for novelty and additional analysis, we propose Knowledge-Assisted Data Generation (KADG)\footnote{We define the setup of KADG as zero-shot* to distinguish zero-shot, because KADG is not strictly zero-shot setup due to use of task knowlege.} to include task knowledge in prompts, which aim to close generated texts to the gold distribution and to increase diversity of them.

To address the aforementioned interests, we evaluate \textsc{JapaGen} across various Japanese tasks, including text classification, natural language inference, semantic textual similarity, and linguistic acceptability, in both few-shot and zero-shot learning settings. Furthermore, we propose a novel approach termed Knowledge-Assisted Data Generation (KADG)\footnote{We define the setup of KADG as zero-shot* to distinguish it from strict zero-shot methods due to the incorporation of task knowledge.}, which integrates task-specific knowledge into prompts to align generated texts more closely with gold-standard distributions and enhance text diversity (\S\ref{sec:kadg}).

%Our experimental results demonstrate that both few- and zero-shot \textsc{JapaGen} obtains higher or near performances compared to few-shot prompting and few-shot fine-tuning in some tasks. Additional analysis shows that KADG make generated texts closer to the gold distribution while keeping label correctness, but not always increase performances.

%Our experimental results demonstrate that both few-shot and zero-shot \textsc{JapaGen} consistently achieve competitive or superior performance compared to traditional few-shot prompting and fine-tuning methods across several tasks. Furthermore, our analysis reveals that KADG improves the fidelity of generated texts to gold distributions while maintaining label accuracy, although it does not universally enhance overall task performance.

Our experiments indicate that, in five out of six tasks, zero-shot \textsc{JapaGen} outperforms few-shot BERT fine-tuning.
Moreover, \textsc{JapaGen} demonstrates superior performances in two tasks compared to few-shot \textsc{Prompting}.
These experimental results suggest that \textsc{JapaGen} has the potential to surpass settings with more parameters and more annotated data.
Additionally, our analysis shows that KADG enhances the fidelity of generated texts to gold-standard distributions while maintaining label accuracy, although it does not consistently improve overall task performance.

In summary, our contributions are four-fold:
\begin{enumerate}
    \item We empirically evaluate \textsc{JapaGen}, leveraging LLMs as synthetic data generators, across various Japanese NLP tasks.
    \item We demonstrate the effectiveness of \textsc{JapaGen}, particularly in classification tasks with formal text inputs.
    \item We analyze the impact of dataset size on \textsc{JapaGen}, observing performance improvements with larger synthetic datasets that eventually reach saturation.
    \item We propose and evaluate KADG, demonstrating its potential to refine synthetic data distributions to align with gold standards, thereby enhancing the robustness of \textsc{JapaGen}.
\end{enumerate}
%\item Our code is available \url{https://github.com/retrieva/LLM-data-generation}.

%近年，大規模言語モデル (Large Language Model;LLM) に関する研究が盛んに行われている．Brownら [1] は GPT-3 を開発し，LLM にタスクの説明文と数件の教師データを入力することで，LLM のパラメータ更新を行うことなく高い精度を達成した．日本語においても，同様に高い性能を達成できる場合もあることが報告されている1）．このように LLM は従来の機械学習モデルと比べて優れた特徴を持つ一方で，パラメータ数が多く，推論時には GPU リソースの制約などの運用コストが高い．

\section{Related Work}
\subsection{Efficient Learning Strategies with LLMs} \label{sec:prompting}
Large Language Models (LLMs) exhibit high performance across various tasks using few-shot or zero-shot learning paradigms. Despite their capabilities, LLMs have numerous parameters, leading to substantial operational costs. To address these challenges, several methods for more efficient utilization of LLMs have been proposed. One such method is {\textsc{Prompting}}, which enables LLMs to perform tasks effectively without requiring parameter updates. This is achieved by injecting prompts based on task descriptions~\cite{few-shot_learner, gao-etal-2021-making, le-scao-rush-2021-many, zhang2022differentiable}. A prompt consists of input text for the LLM and includes instructions to obtain the desired responses. In few-shot \textsc{Prompting}\footnote{Few-shot \textsc{Prompting} is referred to as In-Context Learning~\cite{few-shot_learner}, however, in this paper, both few-shot and zero-shot \textsc{Prompting} are collectively termed as \textsc{Prompting}.}, the prompt includes a small number of text-label pairs. Compared to traditional fine-tuning, which necessitates costly updates to the LLM's parameters, \textsc{Prompting} improves data efficiency in low-data scenarios. However, Prompting incurs substantial operational costs due to the extensive number of parameters involved.

\subsection{Synthesis of Training Data via LLM}
To reduce the operational costs of LLMs, researchers have recently explored using LLMs as training data generators, followed by fine-tuning smaller task-specific models (TAMs), such as BERT~\cite{devlin-etal-2019-bert}, on the synthetic data. Existing approaches typically employ simple class-conditional prompts and focus on addressing the issues related to the quality of the generated data. Notable early efforts, such as SuperGen~\cite{supergen} and ZeroGen~\cite{zerogen}, have explored the use of LLMs for generating training data for text classification tasks using basic class-conditional prompts. They have also incorporated additional noise-robust learning techniques~\cite{laine2017temporal, wang2019symmetric} to mitigate the quality issues of the generated data. However, it has been reported that balancing the diversity of synthetic datasets with task performance remains challenging~\cite{increasing_diversity}.

To date, these approaches have been primarily validated on English-language tasks. This paper investigates the effectiveness of these methods in mid-resource languages with different linguistic characteristics from English.

\section{Method: \textsc{JapaGen}} \label{sec:japagen}

In this section, we introduce the motivation for synthetic data generation via LLMs in Japanese tasks, define the problem, and describe the methodology for generating synthetic training data for each task. The overview of generating training data via LLMs is illustrated in Figure \ref{fig:overview}.

 %In this section, we first introduce the refusal-aware instruction tuning method (R-Tuning), the core idea of which is divided into two steps: the first step involves identifying and recognizing the uncertain data instances within the instruction tuning dataset, which are beyond the parametric knowledge boundary of the original model. The second step is to construct certain and uncertain dataset. Then, we will detail the instruction tuning and inference extraction process. An illustration of R-Tuning is shown in Figure 2.

% JapaGenはSuperGenの日本語版だよ。日本語で検証する理由は。。。だよ。
% 定義&定式化&説明
% zero-shotとfew-shot
\subsection{Motivation}
%We define \textsc{JapaGen} as \textsc{SuperGen} for Japanese tasks.The reason why we choose Japanese is that Japanese is middle-resource language compared English, and also has different characteristics from English in that no spaces exist between words. It is natural that a powerful LLM can generate high quality pesudo training data, because it is trained mainly English texts and a few texts in other language including Japanese. In this paper, we evaluate \textsc{JapaGen}, \textsc{SuperGen} in Japanese, as a case study of such languages.

We define \textsc{JapaGen} as the Japanese counterpart to \textsc{SuperGen}. The rationale behind selecting Japanese stems from its status as a mid-resource language compared to English, and its different characteristics, such as the absence of spaces between words. Given that powerful LLMs are primarily trained on English texts with limited exposure to other languages including Japanese, it is plausible that they can generate high-quality pseudo training data in English.
In this paper, we evaluate \textsc{JapaGen}, the Japanese version of \textsc{SuperGen}, as a case study focusing on such languages.

\subsection{Problem Definition}

Given the label space $\mathcal{Y}=\{y_i\}_{i=1}^{n}$, we manually create label-descriptive prompts $\mathrm{T}(\text{task},y_i)$.
For prompt details used in our experiments, please refer to \S\ref{sec:prompt}.
We employ LLMs $G_\theta$ to generate training data for encoder models $E_\phi$ (e.g., LSTM~\cite{lstm}, BERT~\cite{devlin-etal-2019-bert}), which are subsequently fine-tuned as estimators.
\textsc{SuperGen} comprises the following three stages:
(1) Synthesizing supervised training data using LLM.
(2) Fine-tuning small models using synthetic data.
(3) Testing the trained model on gold data.

\subsection{Pseudo Data Generation}

In this section, we describe the process of generating pseudo datasets using an LLM for classification and regression tasks.
Our approach includes either a single sentence or a sentence pair as input.
%Please refer to Section \ref{sec:details} for specific details on pseudo dataset generation for each task.

\paragraph{Single Sentence Task} \label{sec:single_gen}
We employ an LLM to generate pseudo-supervised sentences $\tilde{x}_{c,j}$ corresponding to a label $y_c$:
\begin{equation}
  \tilde{x}_{c, j} \sim \textrm{Prob}_{\textrm{LLM}}(\cdot | \textrm{T}(\textrm{task}, y_c)),
  \label{eq:gen}
\end{equation}
where $\textrm{T}(\textrm{task}, y_c)$ represents a prompt including the task description and label $y_c$.
By repeating Equation \ref{eq:gen} $M$ times, we obtain the pseudo dataset $\tilde{D}_{y_c}=\{(\tilde{x}_{c,j},y_c)\}_{j=1}^M$.
Applying this process for all labels $\{y_c\}_{c=1}^C$, we generate the pseudo dataset $\tilde{D}=[\tilde{D}_{y_1},\tilde{D}_{y_2},...,\tilde{D}_{y_C}]$.

\paragraph{Sentence Pair Task}
Initially, we employ an LLM to generate the first sentence $\tilde{x}_{c,j}^1$, analogous to Equation \ref{eq:gen} but excluding the label $y_c$:
\begin{equation}
  \tilde{x}_{c, j}^1 \sim \textrm{Prob}_{\textrm{LLM}}(\cdot | \textrm{T}(\textrm{task})).
  \label{eq:gen_first}
\end{equation}
In the initial phase of sentence generation, the prompt comprises solely the task description.
Subsequently, to generate the second sentence $\tilde{x}_{c,j}^2$, the prompt is augmented to include the task description, the first sentence $\tilde{x}_{c,j}^1$, and the label $y_c$:
\begin{equation}
  \tilde{x}_{c, j}^{2} \sim \textrm{Prob}_{\textrm{LLM}}(\cdot | \textrm{T}(\textrm{task}), \textrm{T}(\textrm{task}, \tilde{x}_{c, j}^{1}, y_c)).
  \label{eq:gen_second}
\end{equation}
By repeating Equations \ref{eq:gen_first} and \ref{eq:gen_second} $M$ times, we generate the pseudo dataset $\tilde{D}_{y_c}=\{(\tilde{x}_{c,j}^1,\tilde{x}_{c,j}^2,y_c)\}_{j=1}^M$.
Applying this process for all labels $\{y_c\}_{c=1}^C$, we obtain the pseudo dataset $\tilde{D}=[\tilde{D}_{y_1},\tilde{D}_{y_2},...,\tilde{D}_{y_C}]$.

\subsection{Knowledge-Assisted Data Generation} \label{sec:kadg}
The diversity of synthetic datasets significantly enhances dataset quality, a critical factor in improving task performance~\cite{increasing_diversity}.
Previous studies attempted to diversify text generation by adjusting hyperparameters such as Top-p and temperature.
However, this approach may compromise label accuracy.
In this paper, we introduce \textit{Knowledge-Assisted Data Generation} (KADG) to enhance dataset diversity while maintaining label correctness.

For each task, we manually create a set of task-specific words $S_{\textrm{task}}$, and randomly select a word $d$ from this set.
We construct a prompt based on the task description, label $y_c$, and the selected task-specific word $d$:
\begin{align}
  d &\sim S_{\textrm{task}}, \\
  \tilde{x}_{c,j} &\sim \textrm{Prob}_{\textrm{LLM}}(\cdot | \textrm{T}(\textrm{task}, y_c, d)).\label{eq:KADG}
\end{align}
By following a process similar to Section \ref{sec:single_gen} across all classes, we generate the synthetic dataset $\tilde{D}$.
For the actual prompts used in our experiments, please refer to \S\ref{sec:prompt}.

%前述の疑似データ生成手法は式 (1) に基づいて，Top-p などの生成パラメータによって調整した分布から生成している．我々の事前実験では，Ye ら [2]と同様の方法で多様な日本語テキストの生成を試みたが，いくつかのタスクで類似したテキストが多く生成された (§4.2)．そこで，本研究ではタスクの知識を Prompt に含めることによって多様なテキストの生成を試みる．タスクごとにそのタスクに関する内容語集合 ��taskを人手で作成し，その中から内容語 �� ∈ ��task をランダムに選択して Prompt を作成する．�� ∼ ��task (5)��˜��, �� ∼ ProbLLM(·|T(task, ����, ��)) (6)T(task, ����, ��) はタスク task の説明文と，クラス ���� の説明文，内容語 �� に関連する Prompt を生成している．この処理を §3.1 と同様に全てのクラスに対して実施し，疑似データセット ��˜ を作成する．内容語を考慮した疑似データ生成方法を Knowledge-AssistedData Generation (KADG) と呼ぶ．KADG は教師データを使用しないという点では Zero-Shot と見なすこともできるが，タスク知識を利用しているため，これまでの Zero-Shot と同一と述べるのは適当ではないと考える．そこで本研究では Zero-Shot に対してKADG を適応した場合の問題設定を Zero-Shot* と呼称し，Zero-Shot と区別する．

\section{Experiment}

In this section, we present an overview of the benchmark datasets, the corresponding evaluation settings, the baseline methods, and the implementation details.
Subsequently, we compare our \textsc{JapaGen} to baseline methods in both few-shot and zero-shot settings.

\subsection{Setup}
\paragraph{Benchmarks.}
%To evaluate performance of \textsc{JapaGen} on various tasks, we used following benchmarks: MARC-ja, JSTS, JNLI, and JCoLA from JGLUE~\cite{kurihara-etal-2022-jglue}. To test on diverse domain, we also used two datasets for news topic classification (News) and SNS fact classification (COVID-19). All of them are Japanese tasks. JSTS is sentence similarity estimation task, and the others are text classification tasks. We adopt Spearman score as the metrics for JSTS, Matthews Correlation Coefficient (MCC;~\cite{mcc}) for JCoLA, and Accuracy for the other tasks. More details (\textit{e.g.}, dataset statistics, task explanation) are described in \S\ref{sec:data_task}.

To evaluate \textsc{JapaGen} across various tasks, we used the following benchmarks from JGLUE~\cite{kurihara-etal-2022-jglue}: MARC-ja, JSTS, JNLI, and JCoLA. Additionally, to test across diverse domains, we also used two datasets for news topic classification (News) and SNS fact classification (COVID-19).
All of these benchmarks are Japanese tasks. JSTS involves sentence similarity estimation, while the others are text classification tasks. We evaluated using Spearman's rank correlation coefficient (Spearman score) for JSTS, Matthews correlation coefficient (MCC;~\cite{mcc}) for JCoLA, and Accuracy for the remaining tasks.
For more detailed information such as dataset statistics and task explanations, please refer to Section \ref{sec:data_task}.

\begin{table*}[t]
\begin{center}
\vspace{-2mm}
\small
\resizebox{\linewidth}{!}{%
\begin{tabular}{lccccccc}
\toprule
\textbf{Method}& \textbf{MARC-ja} & \textbf{JSTS} & \textbf{JNLI} & \textbf{JCoLA} & \textbf{News} & \textbf{COVID-19} & \textbf{Avg.} \\
& Acc. & Spearman & Acc. & Mcc. & Acc. & Acc. & \\
\midrule
\multicolumn{8}{l}{\textit{\textbf{\textsc{Fine-Tuning}:} fine-tuning pretrained BERT under gold data.}}\\
\midrule
Fully Supervised & 95.78\tiny{$\pm$0.1} & 87.47\tiny{$\pm$0.5} & 90.19\tiny{$\pm$0.4} & 40.62\tiny{$\pm$1.2} & 95.75\tiny{$\pm$0.4} & 78.49\tiny{$\pm$0.3} & 82.82 \\
Few-Shot & 61.57\tiny{$\pm$8.5} & 14.80\tiny{$\pm$11.3} & 37.72\tiny{$\pm$13.4} & -0.85\tiny{$\pm$3.5} & 51.98\tiny{$\pm$5.3} & 42.24\tiny{$\pm$9.4} & 37.40 \\
\midrule
\multicolumn{8}{l}{\textit{\textbf{\textsc{Prompting}:} prompt-based LLM learning.}}\\
\midrule
Zero-Shot & 94.82\tiny{$\pm$0.2} & 68.53\tiny{$\pm$0.6} & 41.53\tiny{$\pm$1.0} & 24.76\tiny{$\pm$1.2} & 40.27\tiny{$\pm$1.3} & 62.76\tiny{$\pm$0.6} & 57.66 \\
Few-Shot & 97.38\tiny{$\pm$0.2} & 78.50\tiny{$\pm$2.0} & 35.86\tiny{$\pm$5.3} & 26.00\tiny{$\pm$2.9} & 44.82\tiny{$\pm$2.9} & 65.44\tiny{$\pm$3.4} & 61.72 \\
\midrule
\multicolumn{8}{l}{\textit{\textbf{\textsc{JapaGen}:} fine-tuning pretrained BERT under pseudo training data generated via LLM.}}\\
\midrule
Zero-Shot & 77.76\tiny{$\pm$5.4} & \cellcolor[rgb]{0.9, 0.9, 0.9}\textbf{72.47}\tiny{$\pm$0.1} & \cellcolor[rgb]{0.9, 0.9, 0.9}\textbf{46.49}\tiny{$\pm$1.5} & 18.17\tiny{$\pm$1.7} & \cellcolor[rgb]{0.9, 0.9, 0.9}\textbf{57.37}\tiny{$\pm$2.1} & 34.36\tiny{$\pm$6.4} & 54.23 \\
\hspace{2mm}w/ KADG & 83.24\tiny{$\pm$6.0} & \cellcolor[rgb]{0.9, 0.9, 0.9}\textbf{71.49}\tiny{$\pm$1.2} & \cellcolor[rgb]{0.9, 0.9, 0.9}\textbf{46.04}\tiny{$\pm$0.4} & 16.22\tiny{$\pm$0.5} & \cellcolor[rgb]{0.9, 0.9, 0.9}\textbf{59.00}\tiny{$\pm$1.4} & 26.29\tiny{$\pm$0.8} & 50.38 \\
Few-Shot & 62.97\tiny{$\pm$7.3} & \cellcolor[rgb]{0.9, 0.9, 0.9}\textbf{72.56}\tiny{$\pm$0.3} & \cellcolor[rgb]{0.9, 0.9, 0.9}\textbf{50.82}\tiny{$\pm$0.8} & 14.54\tiny{$\pm$1.1} & \cellcolor[rgb]{0.9, 0.9, 0.9}\textbf{62.86}\tiny{$\pm$2.8} & 43.13\tiny{$\pm$1.5} &51.15 \\
\bottomrule
\end{tabular}%
}
\end{center}
\caption{Results on six Japanese tasks. Each value is average with standard deviations over five runs. The tasks that \textsc{JapaGen} outperforms zero-shot \textsc{Prompting} are in {\colorbox[rgb]{0.9,0.9,0.9}{\textbf{gray}}}. Zero-shot \textsc{JapaGen} outperforms zero-shot \textsc{Prompting} on JSTS, JNLI, ad News. Few-shot (Only one sample per class) \textsc{JapaGen} can improve performances on JNLI and News.}
\label{table:result}
\end{table*}

\paragraph{Baselines.}
%We compare performances of \textsc{JapaGen} with three baselines: (1) \textsc{Prompting}. Prompt-based learning framework via LLM, as introduced in \S\ref{sec:prompting}. (2) \textsc{Few-Shot Fine-Tuning}. Fine-tuned BERT on five gold samples per class. (3) \textsc{Fuly Supervised}. Fine-tuned BERT on all gold data. We measure performances of \textsc{JapaGen} and \textsc{Prompting} on both zero- and few-shot setting. In their few-shot settings, we take one sample per class, and insert them into prompt.

We compared the performances of \textsc{JapaGen} with three baselines: (1) \textsc{Prompting}, a prompt-based learning framework via LLM, as introduced in Section \ref{sec:prompting}. (2) \textsc{Few-Shot Fine-Tuning}, where BERT is fine-tuned on five gold samples per class. (3) \textsc{Fully Supervised}, where BERT is fine-tuned on all gold data. We evaluated the performances of \textsc{JapaGen} and \textsc{Prompting} in both few- and zero-shot settings. In the few-shot setting, we used one sample per class and incorporated them into the prompt.
To distinguish between the few-shot setting of BERT fine-tuning and the one of \textsc{JapaGen} and \textsc{Prompting}, we refer to the former as "few-shot \raise0.2ex\hbox{\textcircled{\scriptsize{$\mathcal{B}$}}}" and the latter as "few-shot \raise0.2ex\hbox{\textcircled{\scriptsize{$\mathcal{L}$}}}".

\paragraph{Implementation Details.}
%We implemented our experiments using PyTorch~\cite{torch} and Hugging Face Transformers~\cite{huggingface}. For pseudo training data generation, we use OpenAI model, \texttt{gpt-3.5-turbo-0613}\footnote{We use generated texts for not commerce but only study.}. The size of generated data is 25,000 per class. In few-shot setting, we take one sample per class randomly. For bert training, we conduct our experiments on a Single NVIDIA TITAN RTX 24GB GPU, and use \texttt{tohoku-nlp/bert-base-japanese-v3} respectively. For each task, we measure performances over five runs with different random seeds. In few-shot setting, we take five sample per class randomly. More details (\textit{e.g.}, generation parameter, training parameter) are described in \S\ref{sec:details}.

We conducted our experiments using PyTorch~\cite{torch} and Hugging Face Transformers~\cite{huggingface}. For synthetic data generation, we utilized the OpenAI model \texttt{gpt-3.5-turbo-0613}\footnote{The generated texts are used solely for study purposes, not for commercial use.}. The size of the generated data was 25,000 per class. In the few-shot setting \raise0.2ex\hbox{\textcircled{\scriptsize{$\mathcal{B}$}}}, one sample per class was randomly selected.
The generation parameters were set to max tokens of 500, top-p of 1.0, temperature of 1.2, and frequency penalty of 0.02, with five pieces of data generated at a time.
In JSTS whose labels are continuous values between 0.0 and 5.0, we set six classes \{0, 1, 2, 3, 4, 5\}.
For the fine-tuning of BERT, we used the pretrained BERT\footnote{\href{https://huggingface.co/tohoku-nlp/bert-base-japanese-v3}{\texttt{tohoku-nlp/bert-base-japanese-v3}}} and performed our experiments on a single NVIDIA TITAN RTX 24GB GPU.
The training parameters\footnote{We set training parameters based on \cite{kurihara-etal-2022-jglue}.} were set to batch size of 32, epoch of 4, label smooth temperature of 0.1, optimizer of AdamW with learning rate of 5e-5, $\beta_1$ of 0.9, $\beta_2$ of 0.999, warmup ratio of 0.1.
Additionally, we set max token length of 512, 512, 512, 128, 512, 384 for MARC-ja, JNLI, JSTS, JCoLA, News, and COVID-19 respectively.
For each task, we measured performances over five runs with different random seeds. In the few-shot setting \raise0.2ex\hbox{\textcircled{\scriptsize{$\mathcal{L}$}}}, we randomly selected five samples per class.

\begin{figure*}
\includegraphics[width=\linewidth]{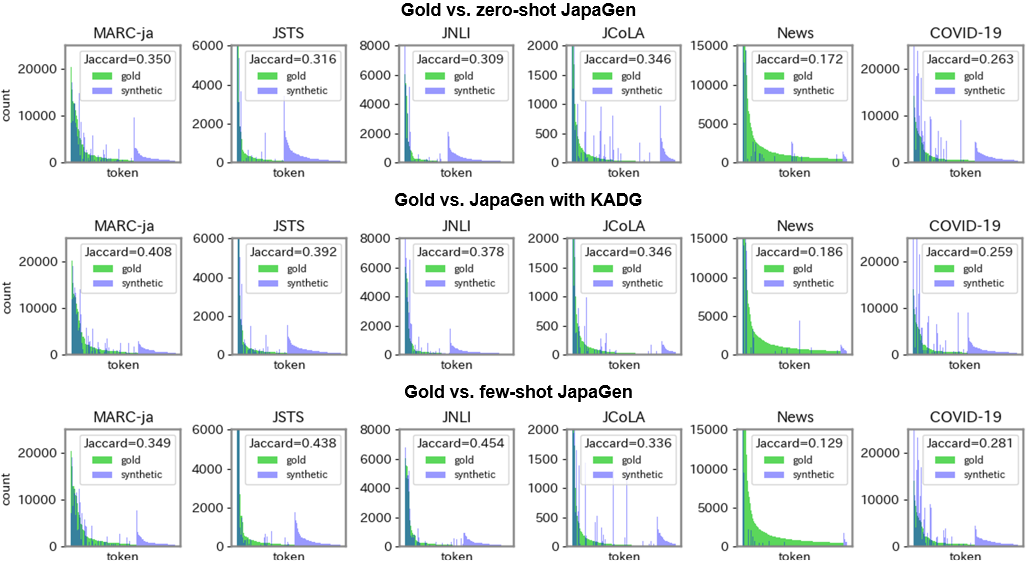}
   \caption{Distribution of the number of appeared tokens between gold and synthetic dataset. Top: zero-shot \textsc{JapaGen}, Middle: \textsc{JapaGen} with KADG, and Bottom: few-shot \textsc{JapaGen}. Compared to zero-shot \textsc{JapaGen}, KADG can improve alignment between gold and synthetic dataset on MARC-ja, JSTS, JNLI, and News. Few-shot \textsc{JapaGen} can also improve alignment on JSTS, JNLI, and COVID-19.}
\label{fig:worddup}
\end{figure*}

%For more details such as generation parameters and training parameters, please refer to Section \ref{sec:details}

\subsection{Experimental Results} \label{sec:result}

In this section, we compare \textsc{JapaGen} to baselines.
Our experimental results are shown in Table \ref{table:result}.

\subsubsection*{Zero-shot \textsc{JapaGen} vs. \textsc{Fine-Tuning}}
%First, we compare \textsc{JapaGen} to \textsc{Fine-Tuning} on gold data whose setting is the same model size but larger number of gold dataset.

%Compared to zero-shot \textsc{JapGen}, the setting of BERT fine-tuned on gold data is the same model size but larger number of gold data. It is well-known that the zero-shot way cannot outperform task-specific trained model on human annotated data. In Table\ref{table:result}, \textsc{JapaGen} is no exception to the aforementioned, and so underperforms fully supervised fine-tuning on all tasks. However, \textsc{JapaGen} outperforms few-shot fine-tuning on five tasks except COVID-19. Particularly, on JSTS, \textsc{JapaGen} obtains 57.67\% spearman score higher than few-shot fine-tuning. This result means that \textsc{JapaGen} can be effective in the case that the cost of data-collection or annotation is expensive.

Compared to zero-shot \textsc{JapaGen}, BERT fine-tuned on gold data uses the same model size but with a larger amount of annotated data. It is well-known that the zero-shot approach cannot outperform task-specific models trained on human-annotated data.
In Table \ref{table:result}, \textsc{JapaGen} adheres to this rule, underperforming compared to fully supervised fine-tuning across all tasks. However, \textsc{JapaGen} outperforms few-shot fine-tuning on five tasks except for COVID-19. Notably in JSTS, \textsc{JapaGen} achieves a Spearman score of 57.67\%, exceeding the performance of few-shot \raise0.2ex\hbox{\textcircled{\scriptsize{$\mathcal{B}$}}} fine-tuning.
This result suggests that \textsc{JapaGen} can be effective in scenarios where the cost of data collection or annotation is high.

\subsubsection*{Zero-shot \textsc{JapaGen} vs. \textsc{Prompting}}
%Second, we compare \textsc{JapaGen} to \textsc{Prompting} whose setting is the significantly larger model size but the same or slightly larger gold dataset size.
%Compared to zero-shot \textsc{JapaGen}, the setting of \textsc{Prompting} is significantly larger model size but the same or slightly larger number of gold data in zero- or few-shot setup. In Table\ref{table:result}, \textsc{JapaGen} achieves 3.94\%, 4.96\%, and 17.10\% higher than zero-shot \textsc{Prompting} on JSTS, JNLI, and News respectively. These tasks commonly have formal text as input. Note that, \textsc{JapaGen} also outperforms few-shot \textsc{Prompting} on JNLI and News, which means that \textsc{JapaGen} has potential to outperform the strong setup of more parameters and more gold data. These tasks are commonly classification tasks as well as have formal text as input.

Compared to zero-shot \textsc{JapaGen}, \textsc{Prompting} employs a significantly larger model size.
%but the same or slightly larger amount of annotated data in zero- or few-shot setups.
In Table \ref{table:result}, \textsc{JapaGen} achieves performance improvements of 3.94\%, 4.96\%, and 17.10\% over zero-shot \textsc{Prompting} on JSTS, JNLI, and News, respectively. These tasks typically involve formal text as input.
Moreover, \textsc{JapaGen} also surpasses few-shot \raise0.2ex\hbox{\textcircled{\scriptsize{$\mathcal{L}$}}} \textsc{Prompting} on JNLI and News, suggesting that \textsc{JapaGen} has the potential to outperform settings with more parameters and more annotated data. These tasks are commonly classification tasks that involve formal text as input.

\subsubsection*{KADG and \textsc{JapaGen}}
%We attempt to inject task knowledge into prompt to improve the performance of \textsc{JapaGen}, because prompt engineering can enhance LLM's ability, and improve generated text quality~\cite{prompt_engineering1, prompt_engineering2, prompt_engineering3}. In Table\ref{table:result}, KADG outperforms \textsc{JapaGen} only on MARC-ja and News, but not improves performances on the other four tasks. Especially, on MARC-ja, \textsc{JapaGen} obtains 6.52\% higher than KADG. Therefore, prompt engineering might be effective in specific tasks. Few-shot setup outperforms zero-shot setup on JSTS, JNLI, News, and COVID-19. Especially, few-shot setup obtains 4.33\%, 5.49\%, and 8.77\% higher than zero-shot setup. Injecting task knowledge into prompt or using few-shot sample can bring generated texts closer to gold texts, but restrict diversity of synthetic dataset. Detailed analysis is described in \S\ref{sec:analysis}.

We attempt to enhance the performance of \textsc{JapaGen} by injecting task knowledge into prompts, as prompt engineering has been shown to enhance the capability of LLMs and improve the quality of generated text~\cite{prompt_engineering1, prompt_engineering2, prompt_engineering3}.
In Table \ref{table:result}, KADG outperforms zero-shot\textsc{JapaGen} only on MARC-ja and News, but does not improve performance on the other four tasks. Specifically, KADG achieves a 5.48\% higher score than \textsc{JapaGen} on MARC-ja. This suggests that prompt engineering may be particularly effective for specific tasks.
In \textsc{JapaGen}, the few-shot \raise0.2ex\hbox{\textcircled{\scriptsize{$\mathcal{L}$}}} setting consistently outperforms the zero-shot setting on JSTS, JNLI, News, and COVID-19. Notably, the few-shot setting achieves improvements of 4.33\%, 5.49\%, and 8.77\% over the zero-shot settings on JNLI, News, and COVID-19, respectively.
Injecting task knowledge into prompts or using few-shot samples can bring generated texts closer to gold-standard texts, but it may restrict the diversity of the synthetic dataset. A detailed analysis is provided in \S\ref{sec:analysis}.

\subsection{Additional Analysis} \label{sec:analysis}

In this section, we analyze \textsc{JapaGen} on distribution, diversity, and label correctness of synthetic and gold datasets.
Then, we qualitatively evaluate synthetic data for each task.

\paragraph{Distribution.}
%One of the important factors for task performances is the alignment between gold data and synthetic data distribution. Figure\ref{fig:worddup} represents the distribution of the number of tokens appeared in dataset. We also measure the weighted jaccard index to quantitatively evaluate the degree of alignment between gold and synthetic distributions. We use 1,000 samples per class for distribution analysis. The top and middle of Figure\ref{fig:worddup} show that KADG can obtain higher jaccard index than zero-shot \textsc{JapaGen} on MARC-ja, JSTS, JNLI, and News. The top and bottom of Figure\ref{fig:worddup} shows that few-shot \textsc{JapaGen} can obtain higher jaccard index than zero-shot one on JSTS, JNLI, and News. Qualitatively, we observe that the number of words appear only in synthetic dataset decreases by KADG and few-shot setup. This results suggest that designing prompt and using few real samples may bring synthetic data distribution closer to gold one.

One of the critical factors influencing task performance is the alignment between the distributions of gold data and synthetic data.
To observe this alignment, we compare token appearances within their respective datasets in a simple manner.
Figure \ref{fig:worddup} represents the distribution of token frequencies within the dataset. We also quantitatively assess the alignment using the weighted Jaccard index, based on 1,000 samples per class for distribution analysis.
In the top and middle sections of Figure \ref{fig:worddup}, KADG achieves a higher Jaccard index compared to zero-shot \textsc{JapaGen} for MARC-ja, JSTS, JNLI, and News. Conversely, in the top and bottom sections of Figure \ref{fig:worddup}, few-shot \textsc{JapaGen} outperforms zero-shot \textsc{JapaGen} regarding the Jaccard index for JSTS, JNLI, and News. Qualitatively, we observe a decrease in the number of words appearing only in the synthetic dataset, the blue-only part in Figure \ref{fig:worddup}, with KADG and the few-shot setting.
These results suggest that designing effective prompts and incorporating a few real samples can help bring the synthetic data distribution closer to that of the gold standard.

\paragraph{Diversity \& Label Correctness.}
%Synthetic dataset may not have diversity, because the same prompt is input into LLM to generate pseudo data. Therefore, we follow previous work~\cite{div_analysis} and analyze diversity of the synthetic dataset and gold ones by measure Self-BLEU~\cite{self-bleu}. This metric means that the lower the value, the more diverse a dataset. Previous studies have reported the trade-off between dataset diversity and label correctness~\cite{increasing_diversity, zerogen}. Therefore, we also analyze label correctness in synthetic dataset. To measure label correctness, we first train BERT on gold training dataset, and then measure accuracy\footnote{In JSTS, we measure Mean Squared Error (MSE).} on synthetic dataset. Table\ref{table:result3} shows the diversity and label correctness on MARC-ja, JSTS, JNLI, and JCoLA due to page limitation\footnote{The diversity on the other dataset refer to \S\ref{sec:additional_analysis}.}. In JSTS, JNLI, and JCoLA, \textsc{JapaGen}'s dataset has the diversity close to gold dataset, and label correctness is not sufficient in JSTS, JNLI, and JCoLA compared to gold dataset.

Synthetic datasets often exhibit limited diversity because they are generated using the same prompt input into the LLM. To assess dataset diversity, we adopt the methodology of a previous study~\cite{div_analysis} and use the Self-BLEU metric~\cite{self-bleu} to compare the diversity of synthetic and gold datasets. A lower Self-BLEU score indicates higher dataset diversity. Previous studies have highlighted a trade-off between dataset diversity and label correctness~\cite{increasing_diversity, zerogen}. Consequently, we also evaluate label correctness in the synthetic dataset. To do so, we first train BERT on the gold training dataset and then measure accuracy\footnote{In JSTS, Mean Squared Error (MSE) is used for measurement.} on the synthetic dataset.
Table \ref{table:result3} presents the diversity and label correctness analysis for each task.

\begin{table}[t]
\begin{center}
\small
\resizebox{\linewidth}{!}{%
\begin{tabular}{lcccc}
\toprule
\textbf{Dataset} & \textbf{MAR.} & \textbf{JSTS}* & \textbf{JNLI} & \textbf{JCoLA} \\
\midrule
\multicolumn{5}{c}{\textbf{\textsc{Diversity}} (\%)} \\
\midrule
Gold & 40.53 & 72.93 & 72.94 & 56.66 \\
Zero-shot & 91.67 & 74.89 & 69.97 & 65.80 \\
w/ KADG & 84.97 & 76.12 & 73.13 & 78.91 \\
Few-shot & 90.25 & 81.80 & 78.28 & 67.15 \\
\midrule
\multicolumn{5}{c}{\textbf{\textsc{Label Correctness}} (\%)} \\
\midrule
Gold & 99.06 & 0.137 & 98.01 & 96.28 \\
Zero-shot & 99.97 & 1.540 & 35.11 & 66.34 \\
w/ KADG & 99.96 & 1.540 & 39.37 & 63.94 \\
Few-shot & 99.90 & 1.094 & 50.16 & 63.33  \\
\bottomrule
\end{tabular}%
}
\end{center}
\caption{Diversity and label correctness of synthetic dataset. We measure the diversity by Self-BLEU. *In JSTS, label correctness is measured by MSE.}
\label{table:result3}
\end{table}

\begin{figure}
\includegraphics[width=\linewidth]{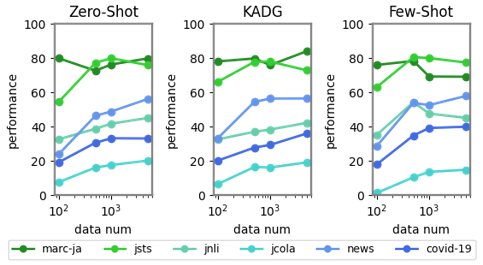}
   \caption{Performance transition with synthetic dataset size on zero-shot, KADG, and few-shot settings.}
\label{fig:data_scaling}
\end{figure}

As shown in the upper part of Table \ref{table:result3}, the Self-BLEU score of the synthetic dataset of zero-shot \textsc{JapaGen} is approximately twice as high, indicating less diversity compared to the gold dataset in MARC-ja.
However, zero-shot \textsc{JapaGen} can synthesize datasets with a diversity similar to the gold dataset in JSTS, JNLI, and JCoLA.
In contrast, in the lower part of Table \ref{table:result3}, the label correctness in JSTS, JNLI, and JCoLA is not as high as in the gold dataset.
Despite reports suggesting that decreasing the Self-BLEU score reduces label accuracy and degrades downstream task performance~\cite{zerogen}, in MARC-ja, KADG improves the Self-BLEU score without compromising label correctness and enhances downstream performance.
The few-shot setting yielded results similar to zero-shot JapaGen in diversity, but improvements in label correctness were observed in the two tasks, JSTS and JNLI.

%\red{TODO: add description for the diversity and label correctness of Few-shot.}

\begin{table*}[t]
\begin{center}
\small
\resizebox{\linewidth}{!}{%
\begin{tabular}{l|l|l}
\toprule
\textbf{Task} & \textbf{Synthesized Text} & \textbf{Label} \\
\midrule\multirow{4}{*}{MARC-ja} & この商品は思っていた以上に素晴らしかったです！購入して本当に良かったです。... & Positive \\
 & (This product was even more nice than I expected! I'm really glad I bought it. ...) & \\
& 商品は非常に不満でした。品質が悪い上に、配送にも遅延がありました。使ってみると... & Negative \\
& (I was extremely dissatisfied with the product. In addition to poor quality, there were delays in delivery. ...) & \\
\midrule
\multirow{4}{*}{JSTS} & 子供たちが講演で楽しそうに遊んでいます。 &  \\
& (The children are having fun playing in the park.) & similarity \\
& 講演で遊ぶ子供たちが笑顔で何かを楽しんでいます。 & \hspace{5mm} = 1.0 \\
& (The children playing in the park are smiling and enjoying something.) &  \\
\midrule
\multirow{12}{*}{JNLI} & 幸せそうなカップルが手をつないで海辺を歩いている。 & \multirow{4}{*}{Entailment} \\
& (A happy couple is walking hand in hand along the seaside.) &  \\
& 青い空と波が背景に広がり、夕日の光が二人を照らしている。 &  \\
& (With the blue sky and waves in the background, the light of the setting sun shines on the couple.) &  \\
\cmidrule(r){2-3}
& 木々が繁茂する森の中で、明るい光が差し込む風景。 & \multirow{4}{*}{Contradiction} \\
& (In the forest where trees grow thickly, bright light streams through the landscape.) &  \\
& 濃い霧がかかり、視界がほとんどない中に立つ孤独な木。 &  \\
& (A solitary tree stands amidst a dense fog, with almost no visibility.) &  \\
\cmidrule(r){2-3}
& 美しい夕焼け空の中、風景画の中に描かれた山々の輪郭が静かに浮かび上がっている。 & \multirow{4}{*}{Neutral} \\
& (In the beautiful sunset sky, the outlines of mountains depicted in the landscape painting quietly emerge.) &  \\
& 夕暮れ時に描かれた風景で、美しく彩られた空の中には山々の輪郭が描かれています。 &  \\
& (In the landscape painted at dusk, the outlines of mountains are depicted against a beautifully colored sky.) &  \\
\midrule
\multirow{4}{*}{JCoLA} & 私は友達と昨日\uwave{食べった}寿司にします。 & Unacceptable \\
& (I will have the sushi I \uwave{atte} with my friends yesterday.) &\\
& 昨日の夜、友達とおいしいラーメンを食べました。 & Acceptable \\
& (Last night, I ate delicious ramen with my friends.) &\\
\midrule
\multirow{8}{*}{COVID-19} & COVID-19の最新情報です。感染拡大を防ぐためには、手洗いやマスクの着用、人との距離... & General Fact \\
& (Here is the latest information on COVID-19. To prevent the spread of infection, it is important,...) &\\
 & 今日は友人がCOVID-19に感染していました。心配ですが、早く回復することを... & Personal Fact \\
& (Today, my friend tested positive for COVID-19. I'm worried, but I hope they recover quickly...) &\\
 & 新型コロナウイルスの感染が拡大する中、マスクの着用や手洗いの重要性を再認識し... & Opinion \\
& (Amid the spread of the novel coronavirus, I have come to realize once again the importance...) &\\
 & 今日はおいしいお寿司を食べました！旬のネタが特に美味しかったです！ & Impression \\
& (Today, I had some delicious sushi! The seasonal toppings were especially tasty!) &\\
\midrule
\multirow{8}{*}{News} & 日本の低価格航空会社PeachAviationは、ユーザーにより快適なフライト体験を提供する & \multirow{2}{*}{Peachy} \\
& ための新しい取り組みを発表しました。 &  \\
& (Japan's low-cost airline Peach Aviation has announced a new initiative to provide users &\\
& with a more comfortable flight experience.) &\\
& 日本の航空会社、エスマックスが業績好調であることが報じられました。新たな路線の & \multirow{2}{*}{S-MAX} \\
& 開設や購入した新型機の稼働により、利益が大幅に上昇しています。&  \\
& (It has been reported that Japan's airline, Smax, is experiencing strong performance. The opening of &\\
& new routes and the operation of newly purchased aircraft have significantly increased their profits.) &\\
\bottomrule
\end{tabular}%
}
\end{center}
\caption{Synthesized data sample by zero-shot \textsc{JapaGen} for each task.}
\label{table:qualitative}
\end{table*}

\paragraph{Data Scaling.}
%We analyze performances on synthetic data scaling. Figure\ref{fig:data_scaling} shows that, on almost tasks, the larger synthetic dataset size, the higher performances. However, the performance saturates because the performances on 5,000 samples are similar to those on 50,000.

We analyze the performance scaling with respect to data size.
Figure~\ref{fig:data_scaling} demonstrates that for most tasks, performance improves as the data size  increases.
However, performance tends to plateau, as the results with 5,000 samples are similar to those with 50,000 samples.

\subsection{Qualitative Evaluations} \label{sec:qualitative}
%In JCoLA, we find that LLM is poor at generating ungrammatical sentences. In News, we find that synthetic texts are not match to a label if the label is not commonsense such as a proper noun. In JSTS, the label is continuous value, but we can give discrete value into prompt. Therefore, we cannot give detailed similarity between two sentences into prompt.

We observe that \textsc{JapaGen} was generally able to synthesize texts in accordance with the tasks. Below, we describe examples where \textsc{JapaGen} did not perform well for each task.

\paragraph{MARC-ja.}
\textsc{JapaGen} tends to generate similar texts such as "この商品は良い/悪いです。(This commodity is good/bad.)". Table \ref{table:result3} also indicates a high Self-BLEU score for MARC-ja, implying significant similarity among the synthesized texts.
As indicated by the high score of label correctness in Table \ref{table:qualitative}, we observe no discrepancy between the synthesized text and the corresponding label.

\paragraph{JSTS.}
While labels are continuous values, employing discrete values as labels in the prompt limits the capability of \textsc{JapaGen} to capture detailed similarity between two sentences.
For instance, the similarity between the two sentences presented in Table \ref{table:qualitative} is 1.0.
However, from the perspective of native Japanese speakers, this similarity should be rated above 3.0.
The label correctness score (MSE) of synthesized texts by \textsc{JapaGen} is also too high, which suggests that several labels are not correct, compared to that of gold data.

\paragraph{JNLI.}
\textsc{JapaGen} exhibits difficulty distinguishing between "Entailment" and "Neutral".
Specifically, text pairs for "Neutral" are frequently misclassified as "Entailment".
The label correctness score (Accuracy) of synthesized texts by \textsc{JapaGen} is also too low compared to that of the gold data.

\paragraph{JCoLA.}
JCoLA is a binary classification task to predict whether a Japanese text is syntactically acceptable or unacceptable.
Our observation indicate that the LLM struggles with generating unacceptable sentences.
Specifically, the expression {\small{"食べった"}} in Table \ref{table:qualitative} is not a syntactic error but a typo.
This is because LLMs are trained to generate syntactically correct sentences, leading to difficulties in generating grammatically incorrect ones.

\paragraph{COVID-19.}
Synthesized texts correspond to each label; however, \textsc{JapaGen} frequently generates similar texts ({{\textit{e.g.},}\small{"手洗い"} (washing hands), \small{"マスク"} (wearing a mask)}) within a label.
The Self-BLEU score of synthetic texts in COVID-19 is much higher, indicating lower diversity compared to gold data presented in Table \ref{table:result3_add}.

\paragraph{News.}
This is a news topic classification task where topic names as labels include entity-like unique expressions.
Synthetic texts frequently fail to align with these labels, particularly when the labels involve proper nouns or lacks common sense.
For instance, in Table \ref{table:qualitative}, "Peachy" is a category indicating news targeting women; however, it generates content about the real airline "Peach (Peach Aviation)".
Similarly, "S-MAX" is a category for software-related news; however, it frequently produces content about fictional people or companies named 'S-MAX' are often generated.

Throughout all six tasks, while the text synthesized by \textsc{JapaGen} has challenges in terms of diversity and label consistency, it was generally able to produce text that aligned with the tasks.

\subsection{Overall Results}
%In this section, we summarize sections A, B, and C related to the experimental results and analysis. The results of zero-shot JapaGen, when compared to few-shot fine-tuning and prompting, showed that it is particularly effective for classification tasks with formal text input. This suggests that JapaGen has the potential to surpass scenarios with more parameters and more annotations. Additionally, the results from KADG and few-shot JapaGen indicated that incorporating task knowledge and examples into the prompts can further enhance its capabilities. On the other hand, challenges include low label accuracy and the difficulty in synthesizing continuous value labels and obtaining the desired grammatical errors in JCoLA.
In this section, we summarize \S\ref{sec:result}, \S\ref{sec:analysis}, and \S\ref{sec:qualitative} related to the experimental results and analysis. The results of zero-shot \textsc{JapaGen}, comparing to few-shot fine-tuning and prompting, showed that it is particularly effective for classification tasks with formal text input. This suggests \textsc{JapaGen} has the potential to surpass scenarios with more parameters and more annotations. Additionally, the results from KADG and few-shot \textsc{JapaGen} indicated that incorporating task knowledge and examples into the prompts can further enhance its capabilities. On the other hand, challenges include low label correctness and the difficulty in synthesizing datasets with continuous value labels such as JSTS and with the desired grammatical errors in JCoLA.

\section{Conclusion}
%To investigate that \textsc{SuperGen} works well in middle-resource language that has different characteristics from English, we test \textsc{SuperGen} in Japanese tasks, \textsc{JapaGen}.  Our experimental results shows that \textsc{JapaGen} is effective on classification tasks whose input is formal text.

To investigate the effectiveness of \textsc{SuperGen} in a mid-resource language with characteristics different from English, we evaluated \textsc{SuperGen} specifically for Japanese tasks, termed \textsc{JapaGen}. Our experimental results demonstrate that \textsc{JapaGen} is particularly effective for classification tasks where the input consists of formal text compared to few-shot \textsc{Prompting}.

\section*{Future Work}
\begin{itemize}
    \item We will examine the efficacy of prompts in synthesizing high-quality texts for specific tasks.
    \item As the development of open LLMs is also progressing rapidly, we would like to evaluate \textsc{JapaGen} using such LLMs.
\end{itemize}

\section*{Limitation}
\begin{itemize}
    \item Our trained models are unavailable for commercial use because we used OpenAI LLM for data generation.
    \item Although we used GPT-3.5 as a pseudo training data generator, using more advanced LLM (\textit{e.g.}, GPT-4) might yield different results.
    \item To examine the impact of \textsc{SuperGen} on languages with distinct characteristics from English and classified as mid-resource, we selected Japanese as a case study.
    Future research will address additional languages.
\end{itemize}

\section*{Ethics Statement}
While PLMs have demonstrated remarkable capabilities in text generation and comprehension, they also pose potential risks or harms~\cite{bender-koller-2020-climbing, bender-gebru}, such as generating misinformation~\cite{pagnoni-etal-2021-understanding} or amplifying harmful biases~\cite{prabhumoye-etal-2018-style}.
Our work specifically focuses on leveraging existing PLMs to generate training data for NLU tasks, rather than on developing new PLMs or generation methods.
In this study, we comply with the OpenAI's terms of use by not disclosing synthetic data and by refraining from using it for purposes other than study.
Furthermore, this study did not involve any sensitive data but only used publicly available data, including MARC-ja, JSTS, JNLI, JCoLA, News, and COVID-19.

%\section*{Acknowledgement}
%This work is supported by the New Energy and Industrial Technology Development Organization (NEDO) Program No. JPNP18002.

\clearpage

\bibliography{anthology,custom}

\begin{thebibliography}{35}
\expandafter\ifx\csname natexlab\endcsname\relax\def\natexlab#1{#1}\fi

\bibitem[{Bender et~al.(2021)Bender, Gebru, McMillan-Major, and Shmitchell}]{bender-gebru}
Emily~M. Bender, Timnit Gebru, Angelina McMillan-Major, and Shmargaret Shmitchell. 2021.
\newblock On the dangers of stochastic parrots: Can language models be too big?
\newblock In \emph{Proceedings of the 2021 ACM Conference on Fairness, Accountability, and Transparency}, page 610–623.

\bibitem[{Bender and Koller(2020)}]{bender-koller-2020-climbing}
Emily~M. Bender and Alexander Koller. 2020.
\newblock \href {https://doi.org/10.18653/v1/2020.acl-main.463} {Climbing towards {NLU}: {On} meaning, form, and understanding in the age of data}.
\newblock In \emph{Proceedings of the 58th Annual Meeting of the Association for Computational Linguistics}, pages 5185--5198, Online. Association for Computational Linguistics.

\bibitem[{Bowman et~al.(2015)Bowman, Angeli, Potts, and Manning}]{bowman-etal-2015-large}
Samuel~R. Bowman, Gabor Angeli, Christopher Potts, and Christopher~D. Manning. 2015.
\newblock \href {https://doi.org/10.18653/v1/D15-1075} {A large annotated corpus for learning natural language inference}.
\newblock In \emph{Proceedings of the 2015 Conference on Empirical Methods in Natural Language Processing}, pages 632--642, Lisbon, Portugal. Association for Computational Linguistics.

\bibitem[{Brown et~al.(2020)Brown, Mann, Ryder, Subbiah, Kaplan, Dhariwal, Neelakantan, Shyam, Sastry, Askell, Agarwal, Herbert-Voss, Krueger, Henighan, Child, Ramesh, Ziegler, Wu, Winter, Hesse, Chen, Sigler, Litwin, Gray, Chess, Clark, Berner, McCandlish, Radford, Sutskever, and Amodei}]{few-shot_learner}
Tom Brown, Benjamin Mann, Nick Ryder, Melanie Subbiah, Jared~D Kaplan, Prafulla Dhariwal, Arvind Neelakantan, Pranav Shyam, Girish Sastry, Amanda Askell, Sandhini Agarwal, Ariel Herbert-Voss, Gretchen Krueger, Tom Henighan, Rewon Child, Aditya Ramesh, Daniel Ziegler, Jeffrey Wu, Clemens Winter, Chris Hesse, Mark Chen, Eric Sigler, Mateusz Litwin, Scott Gray, Benjamin Chess, Jack Clark, Christopher Berner, Sam McCandlish, Alec Radford, Ilya Sutskever, and Dario Amodei. 2020.
\newblock \href {https://proceedings.neurips.cc/paper_files/paper/2020/file/1457c0d6bfcb4967418bfb8ac142f64a-Paper.pdf} {Language models are few-shot learners}.
\newblock In \emph{Advances in Neural Information Processing Systems}, volume~33, pages 1877--1901. Curran Associates, Inc.

\bibitem[{Chen et~al.(2015)Chen, Fang, Lin, Vedantam, Gupta, Dollar, and Zitnick}]{chen2015microsoftcococaptionsdata}
Xinlei Chen, Hao Fang, Tsung-Yi Lin, Ramakrishna Vedantam, Saurabh Gupta, Piotr Dollar, and C.~Lawrence Zitnick. 2015.
\newblock Microsoft coco captions: Data collection and evaluation server.

\bibitem[{Chung et~al.(2023{\natexlab{a}})Chung, Kamar, and Amershi}]{chung-etal-2023-increasing}
John Chung, Ece Kamar, and Saleema Amershi. 2023{\natexlab{a}}.
\newblock \href {https://doi.org/10.18653/v1/2023.acl-long.34} {Increasing diversity while maintaining accuracy: Text data generation with large language models and human interventions}.
\newblock In \emph{Proceedings of the 61st Annual Meeting of the Association for Computational Linguistics (Volume 1: Long Papers)}, pages 575--593, Toronto, Canada. Association for Computational Linguistics.

\bibitem[{Chung et~al.(2023{\natexlab{b}})Chung, Kamar, and Amershi}]{increasing_diversity}
John Chung, Ece Kamar, and Saleema Amershi. 2023{\natexlab{b}}.
\newblock \href {https://doi.org/10.18653/v1/2023.acl-long.34} {Increasing diversity while maintaining accuracy: Text data generation with large language models and human interventions}.
\newblock In \emph{Proceedings of the 61st Annual Meeting of the Association for Computational Linguistics (Volume 1: Long Papers)}, pages 575--593, Toronto, Canada. Association for Computational Linguistics.

\bibitem[{Devlin et~al.(2019)Devlin, Chang, Lee, and Toutanova}]{devlin-etal-2019-bert}
Jacob Devlin, Ming-Wei Chang, Kenton Lee, and Kristina Toutanova. 2019.
\newblock \href {https://doi.org/10.18653/v1/N19-1423} {{BERT}: Pre-training of deep bidirectional transformers for language understanding}.
\newblock In \emph{Proceedings of the 2019 Conference of the North {A}merican Chapter of the Association for Computational Linguistics: Human Language Technologies, Volume 1 (Long and Short Papers)}, pages 4171--4186, Minneapolis, Minnesota. Association for Computational Linguistics.

\bibitem[{Gao et~al.(2021)Gao, Fisch, and Chen}]{gao-etal-2021-making}
Tianyu Gao, Adam Fisch, and Danqi Chen. 2021.
\newblock \href {https://doi.org/10.18653/v1/2021.acl-long.295} {Making pre-trained language models better few-shot learners}.
\newblock In \emph{Proceedings of the 59th Annual Meeting of the Association for Computational Linguistics and the 11th International Joint Conference on Natural Language Processing (Volume 1: Long Papers)}, pages 3816--3830, Online. Association for Computational Linguistics.

\bibitem[{He et~al.(2022)He, Zheng, Tay, Gupta, Du, Aribandi, Zhao, Li, Chen, Metzler, Cheng, and Chi}]{prompt_engineering3}
Yun He, Steven Zheng, Yi~Tay, Jai Gupta, Yu~Du, Vamsi Aribandi, Zhe Zhao, Yaguang Li, Zhao Chen, Donald Metzler, Heng-Tze Cheng, and Ed~H. Chi. 2022.
\newblock \href {https://proceedings.mlr.press/v162/he22f.html} {{H}yper{P}rompt: Prompt-based task-conditioning of transformers}.
\newblock In \emph{Proceedings of the 39th International Conference on Machine Learning}, volume 162 of \emph{Proceedings of Machine Learning Research}, pages 8678--8690. PMLR.

\bibitem[{Hochreiter and Schmidhuber(1997)}]{lstm}
Sepp Hochreiter and J\"{u}rgen Schmidhuber. 1997.
\newblock \href {https://doi.org/10.1162/neco.1997.9.8.1735} {Long short-term memory}.
\newblock \emph{Neural Comput.}, 9(8):1735–1780.

\bibitem[{Holtzman et~al.(2020)Holtzman, Buys, Du, Forbes, and Choi}]{div_analysis}
Ari Holtzman, Jan Buys, Li~Du, Maxwell Forbes, and Yejin Choi. 2020.
\newblock \href {https://openreview.net/forum?id=rygGQyrFvH} {The curious case of neural text degeneration}.
\newblock In \emph{International Conference on Learning Representations}.

\bibitem[{Kaplan et~al.(2020)Kaplan, McCandlish, Henighan, Brown, Chess, Child, Gray, Radford, Wu, and Amodei}]{scaling_law}
Jared Kaplan, Sam McCandlish, Tom Henighan, Tom~B. Brown, Benjamin Chess, Rewon Child, Scott Gray, Alec Radford, Jeffrey Wu, and Dario Amodei. 2020.
\newblock \href {http://arxiv.org/abs/2001.08361} {Scaling laws for neural language models}.
\newblock \emph{CoRR}, abs/2001.08361.

\bibitem[{Keung et~al.(2020)Keung, Lu, Szarvas, and Smith}]{keung-etal-2020-multilingual}
Phillip Keung, Yichao Lu, Gy{\"o}rgy Szarvas, and Noah~A. Smith. 2020.
\newblock \href {https://doi.org/10.18653/v1/2020.emnlp-main.369} {The multilingual {A}mazon reviews corpus}.
\newblock In \emph{Proceedings of the 2020 Conference on Empirical Methods in Natural Language Processing (EMNLP)}, pages 4563--4568, Online. Association for Computational Linguistics.

\bibitem[{Kojima et~al.(2022)Kojima, Gu, Reid, Matsuo, and Iwasawa}]{cot}
Takeshi Kojima, Shixiang~(Shane) Gu, Machel Reid, Yutaka Matsuo, and Yusuke Iwasawa. 2022.
\newblock \href {https://proceedings.neurips.cc/paper_files/paper/2022/file/8bb0d291acd4acf06ef112099c16f326-Paper-Conference.pdf} {Large language models are zero-shot reasoners}.
\newblock In \emph{Advances in Neural Information Processing Systems}, volume~35, pages 22199--22213. Curran Associates, Inc.

\bibitem[{Kurihara et~al.(2022)Kurihara, Kawahara, and Shibata}]{kurihara-etal-2022-jglue}
Kentaro Kurihara, Daisuke Kawahara, and Tomohide Shibata. 2022.
\newblock \href {https://aclanthology.org/2022.lrec-1.317} {{JGLUE}: {J}apanese general language understanding evaluation}.
\newblock In \emph{Proceedings of the Thirteenth Language Resources and Evaluation Conference}, pages 2957--2966, Marseille, France. European Language Resources Association.

\bibitem[{Laine and Aila(2017)}]{laine2017temporal}
Samuli Laine and Timo Aila. 2017.
\newblock Temporal ensembling for semi-supervised learning.
\newblock In \emph{International Conference on Learning Representations}.

\bibitem[{Le~Scao and Rush(2021)}]{le-scao-rush-2021-many}
Teven Le~Scao and Alexander Rush. 2021.
\newblock \href {https://doi.org/10.18653/v1/2021.naacl-main.208} {How many data points is a prompt worth?}
\newblock In \emph{Proceedings of the 2021 Conference of the North American Chapter of the Association for Computational Linguistics: Human Language Technologies}, pages 2627--2636, Online. Association for Computational Linguistics.

\bibitem[{Matthews(1975)}]{mcc}
B.W. Matthews. 1975.
\newblock \href {https://doi.org/https://doi.org/10.1016/0005-2795(75)90109-9} {Comparison of the predicted and observed secondary structure of t4 phage lysozyme}.
\newblock \emph{Biochimica et Biophysica Acta (BBA) - Protein Structure}, 405(2):442--451.

\bibitem[{Meng et~al.(2022)Meng, Huang, Zhang, and Han}]{supergen}
Yu~Meng, Jiaxin Huang, Yu~Zhang, and Jiawei Han. 2022.
\newblock \href {https://proceedings.neurips.cc/paper_files/paper/2022/file/0346c148ba1c21c6b4780a961ea141dc-Paper-Conference.pdf} {Generating training data with language models: Towards zero-shot language understanding}.
\newblock In \emph{Advances in Neural Information Processing Systems}, volume~35, pages 462--477. Curran Associates, Inc.

\bibitem[{Miyazaki and Shimizu(2016)}]{miyazaki-shimizu-2016-cross}
Takashi Miyazaki and Nobuyuki Shimizu. 2016.
\newblock \href {https://doi.org/10.18653/v1/P16-1168} {Cross-lingual image caption generation}.
\newblock In \emph{Proceedings of the 54th Annual Meeting of the Association for Computational Linguistics (Volume 1: Long Papers)}, pages 1780--1790, Berlin, Germany. Association for Computational Linguistics.

\bibitem[{OpenAI(2024)}]{gpt4}
OpenAI. 2024.
\newblock \href {http://arxiv.org/abs/2303.08774} {Gpt-4 technical report}.

\bibitem[{Pagnoni et~al.(2021)Pagnoni, Balachandran, and Tsvetkov}]{pagnoni-etal-2021-understanding}
Artidoro Pagnoni, Vidhisha Balachandran, and Yulia Tsvetkov. 2021.
\newblock \href {https://doi.org/10.18653/v1/2021.naacl-main.383} {Understanding factuality in abstractive summarization with {FRANK}: A benchmark for factuality metrics}.
\newblock In \emph{Proceedings of the 2021 Conference of the North American Chapter of the Association for Computational Linguistics: Human Language Technologies}, pages 4812--4829, Online. Association for Computational Linguistics.

\bibitem[{Paszke et~al.(2019)Paszke, Gross, Massa, Lerer, Bradbury, Chanan, Killeen, Lin, Gimelshein, Antiga, Desmaison, Kopf, Yang, DeVito, Raison, Tejani, Chilamkurthy, Steiner, Fang, Bai, and Chintala}]{torch}
Adam Paszke, Sam Gross, Francisco Massa, Adam Lerer, James Bradbury, Gregory Chanan, Trevor Killeen, Zeming Lin, Natalia Gimelshein, Luca Antiga, Alban Desmaison, Andreas Kopf, Edward Yang, Zachary DeVito, Martin Raison, Alykhan Tejani, Sasank Chilamkurthy, Benoit Steiner, Lu~Fang, Junjie Bai, and Soumith Chintala. 2019.
\newblock \href {https://proceedings.neurips.cc/paper_files/paper/2019/file/bdbca288fee7f92f2bfa9f7012727740-Paper.pdf} {Pytorch: An imperative style, high-performance deep learning library}.
\newblock In \emph{Advances in Neural Information Processing Systems}, volume~32. Curran Associates, Inc.

\bibitem[{Prabhumoye et~al.(2018)Prabhumoye, Tsvetkov, Salakhutdinov, and Black}]{prabhumoye-etal-2018-style}
Shrimai Prabhumoye, Yulia Tsvetkov, Ruslan Salakhutdinov, and Alan~W Black. 2018.
\newblock \href {https://doi.org/10.18653/v1/P18-1080} {Style transfer through back-translation}.
\newblock In \emph{Proceedings of the 56th Annual Meeting of the Association for Computational Linguistics (Volume 1: Long Papers)}, pages 866--876, Melbourne, Australia. Association for Computational Linguistics.

\bibitem[{Someya et~al.(2024)Someya, Sugimoto, and Oseki}]{someya-etal-2024-jcola-japanese}
Taiga Someya, Yushi Sugimoto, and Yohei Oseki. 2024.
\newblock {JC}o{LA}: {J}apanese corpus of linguistic acceptability.
\newblock In \emph{Proceedings of the 2024 Joint International Conference on Computational Linguistics, Language Resources and Evaluation (LREC-COLING 2024)}, pages 9477--9488.

\bibitem[{Wang et~al.(2019)Wang, Ma, Chen, Luo, Yi, and Bailey}]{wang2019symmetric}
Yisen Wang, Xingjun Ma, Zaiyi Chen, Yuan Luo, Jinfeng Yi, and James Bailey. 2019.
\newblock Symmetric cross entropy for robust learning with noisy labels.
\newblock In \emph{IEEE International Conference on Computer Vision}.

\bibitem[{Wolf et~al.(2020)Wolf, Debut, Sanh, Chaumond, Delangue, Moi, Cistac, Rault, Louf, Funtowicz, Davison, Shleifer, von Platen, Ma, Jernite, Plu, Xu, Le~Scao, Gugger, Drame, Lhoest, and Rush}]{huggingface}
Thomas Wolf, Lysandre Debut, Victor Sanh, Julien Chaumond, Clement Delangue, Anthony Moi, Pierric Cistac, Tim Rault, Remi Louf, Morgan Funtowicz, Joe Davison, Sam Shleifer, Patrick von Platen, Clara Ma, Yacine Jernite, Julien Plu, Canwen Xu, Teven Le~Scao, Sylvain Gugger, Mariama Drame, Quentin Lhoest, and Alexander Rush. 2020.
\newblock \href {https://doi.org/10.18653/v1/2020.emnlp-demos.6} {Transformers: State-of-the-art natural language processing}.
\newblock In \emph{Proceedings of the 2020 Conference on Empirical Methods in Natural Language Processing: System Demonstrations}, pages 38--45, Online. Association for Computational Linguistics.

\bibitem[{Wu and Hu(2023)}]{prompt_engineering1}
Yangjian Wu and Gang Hu. 2023.
\newblock \href {https://doi.org/10.18653/v1/2023.wmt-1.15} {Exploring prompt engineering with {GPT} language models for document-level machine translation: Insights and findings}.
\newblock In \emph{Proceedings of the Eighth Conference on Machine Translation}, pages 166--169, Singapore. Association for Computational Linguistics.

\bibitem[{Yang et~al.(2023)Yang, Wang, Wang, Quan, Feng, Chen, Khabsa, Wang, Xu, and Liu}]{prompt_engineering2}
Li~Yang, Qifan Wang, Jingang Wang, Xiaojun Quan, Fuli Feng, Yu~Chen, Madian Khabsa, Sinong Wang, Zenglin Xu, and Dongfang Liu. 2023.
\newblock \href {https://doi.org/10.18653/v1/2023.findings-acl.633} {{M}ix{PAVE}: Mix-prompt tuning for few-shot product attribute value extraction}.
\newblock In \emph{Findings of the Association for Computational Linguistics: ACL 2023}, pages 9978--9991, Toronto, Canada. Association for Computational Linguistics.

\bibitem[{Ye et~al.(2022{\natexlab{a}})Ye, Gao, Li, Xu, Feng, Wu, Yu, and Kong}]{zerogen}
Jiacheng Ye, Jiahui Gao, Qintong Li, Hang Xu, Jiangtao Feng, Zhiyong Wu, Tao Yu, and Lingpeng Kong. 2022{\natexlab{a}}.
\newblock \href {https://doi.org/10.18653/v1/2022.emnlp-main.801} {{Z}ero{G}en: Efficient zero-shot learning via dataset generation}.
\newblock In \emph{Proceedings of the 2022 Conference on Empirical Methods in Natural Language Processing}, pages 11653--11669, Abu Dhabi, United Arab Emirates. Association for Computational Linguistics.

\bibitem[{Ye et~al.(2022{\natexlab{b}})Ye, Gao, Wu, Feng, Yu, and Kong}]{progen}
Jiacheng Ye, Jiahui Gao, Zhiyong Wu, Jiangtao Feng, Tao Yu, and Lingpeng Kong. 2022{\natexlab{b}}.
\newblock \href {https://doi.org/10.18653/v1/2022.findings-emnlp.269} {{P}ro{G}en: Progressive zero-shot dataset generation via in-context feedback}.
\newblock In \emph{Findings of the Association for Computational Linguistics: EMNLP 2022}, pages 3671--3683, Abu Dhabi, United Arab Emirates. Association for Computational Linguistics.

\bibitem[{Yu et~al.(2023)Yu, Zhuang, Zhang, Meng, Shen, and Zhang}]{regen}
Yue Yu, Yuchen Zhuang, Rongzhi Zhang, Yu~Meng, Jiaming Shen, and Chao Zhang. 2023.
\newblock \href {https://doi.org/10.18653/v1/2023.findings-acl.748} {{R}e{G}en: Zero-shot text classification via training data generation with progressive dense retrieval}.
\newblock In \emph{Findings of the Association for Computational Linguistics: ACL 2023}, pages 11782--11805, Toronto, Canada. Association for Computational Linguistics.

\bibitem[{Zhang et~al.(2022)Zhang, Li, Chen, Deng, Bi, Tan, Huang, and Chen}]{zhang2022differentiable}
Ningyu Zhang, Luoqiu Li, Xiang Chen, Shumin Deng, Zhen Bi, Chuanqi Tan, Fei Huang, and Huajun Chen. 2022.
\newblock Differentiable prompt makes pre-trained language models better few-shot learners.
\newblock In \emph{International Conference on Learning Representations}.

\bibitem[{Zhu et~al.(2018)Zhu, Lu, Zheng, Guo, Zhang, Wang, and Yu}]{self-bleu}
Yaoming Zhu, Sidi Lu, Lei Zheng, Jiaxian Guo, Weinan Zhang, Jun Wang, and Yong Yu. 2018.
\newblock Texygen: A benchmarking platform for text generation models.
\newblock In \emph{The 41st international ACM SIGIR conference on research \& development in information retrieval}.

\end{thebibliography}
\bibliographystyle{acl_natbib}

\clearpage

\appendix

\section{Appendix} \label{sec:appendix}
\subsection{Dataset and Task} \label{sec:data_task}
We describe the six tasks used in our experiment.
The dataset statistics are presented in Table \ref{tab:dataset_static}.
\paragraph{MARC-ja}
A binary classification task to predict the sentiment of product reviews as positive or negative. The dataset used for this task is derived from the Japanese subset of the Multilingual Amazon Reviews Corpus (MARC)~\cite{keung-etal-2020-multilingual}.
\paragraph{JSTS}
A regression task to predict the semantic similarity score between two sentences. The score ranges from 0 (least similar) to 5 (most similar). The data for this task are sourced from the Japanese version of the MS COCO Caption Dataset~\cite{chen2015microsoftcococaptionsdata} and the YJ Captions Dataset~\cite{miyazaki-shimizu-2016-cross}.
\paragraph{JNLI}
A three-way classification task to predict the relation between two sentences. The possible relations are \{contradiction, neutral, entailment\} reflecting the categories utilized in the Stanford Natural Language Inference (SNLI) dataset~\cite{bowman-etal-2015-large}. The data source for this task is the same as that used for JSTS.
\paragraph{JCoLA}
A binary classification task to predict whether a Japanese text is syntactically acceptable or unacceptable. For further details, please refer to \cite{someya-etal-2024-jcola-japanese}.
\paragraph{News}
A nine-way classification task to predict the news topic of a given news text. The news texts are sourced from \href{https://www.rondhuit.com/download.html#ldcc}{Livedoor News}. The possible topics are \{Trend Topic News, Sports Watch, IT Life hack, Consumer Electronics, MOVIE, DOKUJOTSUSHIN, S-MAX, HOMME, Peachy\}.
\paragraph{COVID-19}
A four-way classification task to predict the factuality of tweets about COVID-19. The categories of factual information include "general fact," "personal fact," "opinion," and "impressions." The data for this task are sourced from \url{https://www.db.info.gifu-u.ac.jp/covid-19-twitter-dataset/}.

\begin{table}[t]
\begin{center}
\small
\resizebox{0.9\linewidth}{!}{%
\begin{tabular}{llccc}
\toprule
\multirow{2}{*}{\textbf{Dataset}} && \multicolumn{3}{c}{\textbf{Number of Samples}}\\
 && Train & Dev. & Test \\
\midrule
\multirow{4}{*}{JGLUE} & MARC-ja & 150,022 & 37,506 & 5,654 \\
&JSTS & 9,960 & 2,491 & 1,457 \\
&JNLI & 16,058 & 4,015 & 2,434 \\
&JCoLA & 4,000 & 1,000 & 865 \\
 \multicolumn{2}{c}{News} & 4,375 & 625 & 1,475 \\
 \multicolumn{2}{c}{COVID-19} & 4,375 & 625 & 7,547 \\
\bottomrule
\end{tabular}%
}
\end{center}
\caption{Dataset statistics.}
\label{tab:dataset_static}
\end{table}

\begin{table}[htbp]
\begin{center}
\small
\resizebox{0.7\linewidth}{!}{%
\begin{tabular}{lcc}
\toprule
\textbf{Dataset} & \textbf{News} & \textbf{COVID-19} \\
\midrule
\multicolumn{3}{c}{\textbf{\textsc{Diversity}} (\%)} \\
\midrule
Gold & 62.97 & 43.14 \\
Zero-shot & 79.90 & 84.31 \\
w/ KADG & 82.93 & 81.91 \\
Few-shot & 79.25 & 83.40 \\
\midrule
\multicolumn{3}{c}{\textbf{\textsc{Label Correctness}} (\%)} \\
\midrule
Gold & 98.89 & 90.87 \\
Zero-shot & 49.84 & 60.80 \\
w/ KADG & 43.61 & 58.86 \\
Few-shot & 57.33 & 64.43 \\
\bottomrule
\end{tabular}%
}
\end{center}
\caption{Diversity and label correctness of synthetic dataset in News and COVID-19.}
\label{table:result3_add}
\end{table}

%\subsection{Additional Analysis} \label{sec:additional_analysis}
%Table \ref{table:result3_add} shows the dataset diversity and label correctness of News and COVID-19, which can not be placed in \S\ref{sec:analysis} due to page limitation.
%\red{TODO: add description about diversity and correctness for NEws and COVID-19.}

\subsection{Metrics} \label{sec:metrics}
\paragraph{Spearman's Correlation Score}
This metric means the consistency between two sets of rankings by calculating the correlation between their ranks. A score close to 1 indicates strong agreement, meaning the model's ranked outputs closely match the true ranked labels.

\paragraph{Matthews Correlation Coefficient (MCC)}
MCC measures the quality of binary classifications by considering true positives, false positives, true negatives, and false negatives in a balanced way. Its value ranges from -1 to 1, where 1 indicates perfect prediction, and -1 a complete inverse relationship.

\paragraph{Self-BLEU}
This metric calculates BLEU scores for generated text samples against other samples within the same set to measure diversity. Lower Self-BLEU indicates more diverse outputs.

\subsection{Additional Results}
The diversity (Self-BLEU) and label correctness of News and COVID-19 are shown in Table \ref{table:result3_add}.
While the diversity of News and COVID-19 in few-shot is lower than that in zero-shot, few-shot \textsc{JapaGen} can improve the label correctness of News and COVID-19.

\subsection{Prompt for Each Task} \label{sec:prompt}
%For prompt details used in our experiments, please refer to \url{https://github.com/retrieva/JapaGen}.
For prompt details used in our experiments, please refer to \url{https://github.com/retrieva/JapaGen} due to the page limitation.

%Table \ref{table:qualitative_add} shows the samples synthesized by \textsc{JapaGen} for each task.
%In MARC-ja, \textsc{JapaGen} tended to generate similar abstract texts like "この商品は良い/悪いです。(This comodity is good/bad.)". Table \ref{table:result3} also show the diversity in MARC-ja is high, which means synthesized texts are similar.
%In JNLI, \textsc{JapaGen} could not understand the difference between "Entailment" and "Neutral". In particular, pair texts for "Neutral" tended to be "Entailment".
%In COVID-19, synthesized texts correspond to each label, but they are similar within a label.

\end{document}